\documentclass[runningheads]{llncs}

\usepackage[mobile]{eccv}

\usepackage{eccvabbrv}

\usepackage{graphicx}
\usepackage{booktabs}

\usepackage[accsupp]{axessibility}  

\usepackage{hyperref}

\usepackage{orcidlink}
\usepackage{multirow}
\usepackage{booktabs}
\usepackage{bbm}

\usepackage{sidecap}
\usepackage{pifont}

\newcommand{\op}[1]{\operatorname{#1}}
\newcommand{\tabincell}[2]{\begin{tabular}{@{}#1@{}}#2\end{tabular}}

\usepackage{caption}

\begin{document}
\makeatletter
\def\thanks#1{\protected@xdef\@thanks{\@thanks\protect\footnotetext{#1}}}
\makeatother

\title{Tracking Meets LoRA: Faster Training, Larger Model, Stronger Performance} 


\author{Liting Lin$^{1}$ \and
Heng Fan$^{2}$ \and
Zhipeng Zhang$^{3}$ \and
Yaowei Wang$^{1,\dagger}$  \and \\
Yong Xu$^{4,1}$ \and
Haibin Ling$^{5,\dagger}$  }

\authorrunning{L. Lin et al.}

\institute{Peng Cheng Laboratory, China \and
Department of CSE, University of North Texas, USA \and
KargoBot, China \and
School of Computer Science \& Engineering, South China Univ. of Tech., China \and
Department of Computer Science, Stony Brook University, USA \\
\email{lt.lin@qq.com, heng.fan@unt.edu, zhipeng.zhang.cv@outlook.com \\
wangyw@pcl.ac.cn, yxu@scut.edu.cn, hling@cs.stonybrook.edu}}

\thanks{$\dagger$ corresponding author}

\maketitle
\begin{abstract}
Motivated by the Parameter-Efficient Fine-Tuning (PEFT) in large language models, we propose \textbf{LoRAT}, a method that unveils the power of larger Vision Transformers (ViT) for tracking within laboratory-level resources. The essence of our work lies in adapting LoRA, a technique that fine-tunes a small subset of model parameters without adding inference latency, to the domain of visual tracking. However, unique challenges and potential domain gaps make this transfer not as easy as the first intuition. Firstly, a transformer-based tracker constructs unshared position embedding for template and search image. This poses a challenge for the transfer of LoRA, usually requiring consistency in the design when applied to the pre-trained backbone, to downstream tasks. Secondly, the inductive bias inherent in convolutional heads diminishes the effectiveness of parameter-efficient fine-tuning in tracking models. To overcome these limitations, we first decouple the position embeddings in transformer-based trackers into shared spatial ones and independent type ones. The shared embeddings, which describe the absolute coordinates of multi-resolution images (namely, the template and search images), are inherited from the pre-trained backbones. In contrast, the independent embeddings indicate the sources of each token and are learned from scratch. Furthermore, we design an anchor-free head solely based on a multilayer perceptron (MLP) to adapt PETR, enabling better performance with less computational overhead. With our design, 1) it becomes practical to train trackers with the ViT-g backbone on GPUs with only memory of 25.8GB (batch size of 16); 2) we reduce the training time of the L-224 variant from 35.0 to 10.8 GPU hours; 3) we improve the LaSOT SUC score from 0.703 to 0.742 with the L-224 variant; 4) we fast the inference speed of the L-224 variant from 52 to 119 FPS. Code and models are available at \url{https://github.com/LitingLin/LoRAT}.
\end{abstract}

\section{Introduction}
\label{sec:intro}

\begin{figure}[t]
\centering
\begin{subfigure}[b]{0.32\textwidth}
    \includegraphics[width=\textwidth]{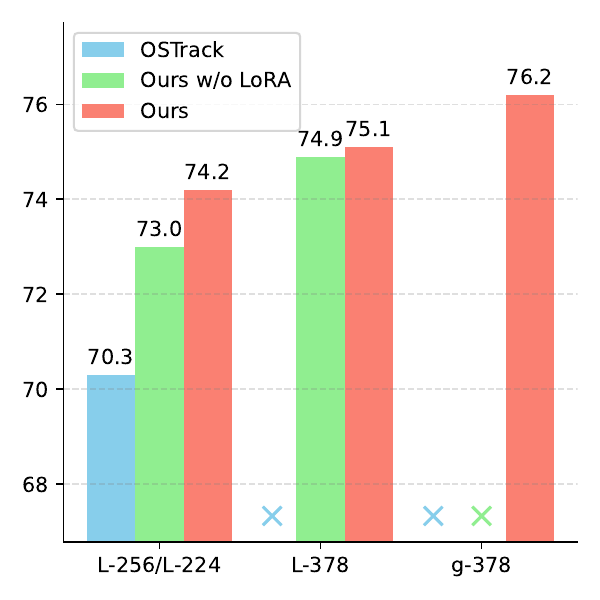}
    \caption{SUC on LaSOT}
\end{subfigure}
\hfill 
\begin{subfigure}[b]{0.32\textwidth}
    \includegraphics[width=\textwidth]{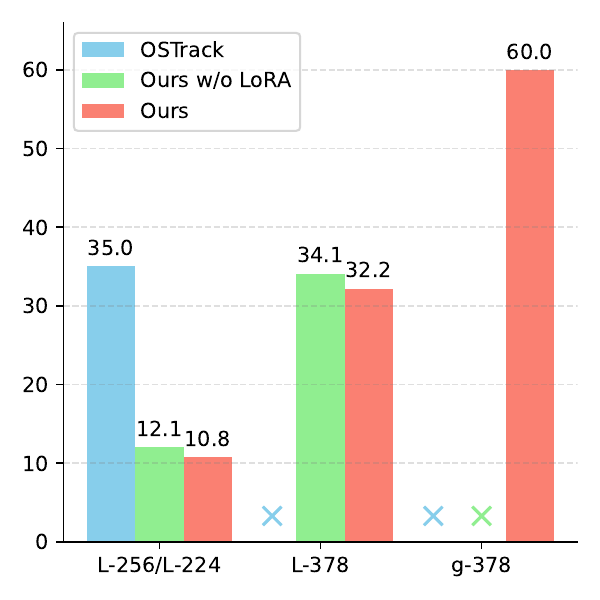}
    \caption{Training Time (h)}
\end{subfigure}
\hfill 
\begin{subfigure}[b]{0.32\textwidth}
    \includegraphics[width=\textwidth]{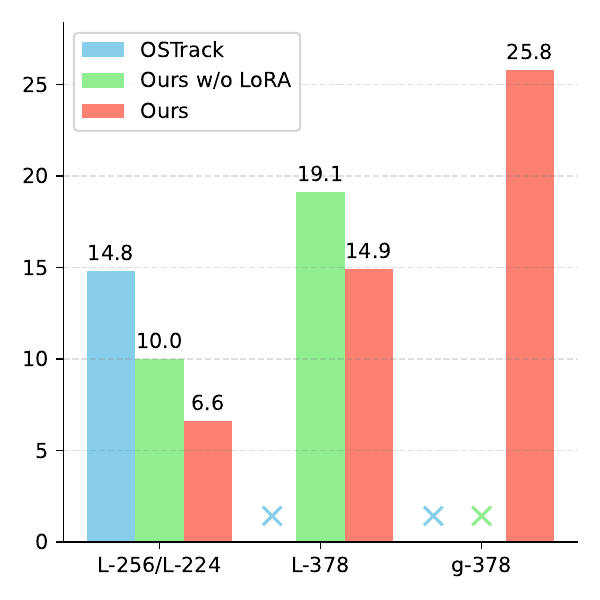}
    \caption{Training Memory (GB)}
\end{subfigure}
\caption{Comparison of tracking models on performance and training efficiency. "$\times$" indicates failure to train due to insufficient memory. Best viewed in color for all figures.}
\label{fig:performance}
\end{figure}

Visual tracking, aiming to continuously locate the target object from a sequence given its initial state, is one of the most fundamental problems in computer vision and has been extensively explored in the past decades~\cite{siamfc,kcf,eco,siamrpnPlus,fan2019siamese,ocean,dimp}. In recent years, the tracking community has witnessed considerable advancements~\cite{transt,stark,swintrack,ostrack} by leveraging the Transformer architecture~\cite{transformer,vit}, primarily owing to its exceptional ability in capturing long-range dependencies and flexibility in accommodating a wide variety of pre-training models~\cite{resnet, swint, vit, mae}. While Transformer contributes to significant improvements in tracking performance, their usage is \emph{resource-intensive}. In particular, most performance-oriented Transformer trackers require extensive computational resources, typically involving multiple high-end data-center GPUs and prolonged training periods. Figure~\ref{fig:performance} illustrates the escalating resource requirements for training large-scale trackers, which are becoming increasingly unaffordable for most researchers. The largest model in visual tracking to date, SeqTrack-L384~\cite{seqtrack}, employs a ViT-L backbone, yet lags behind the scale of available pre-trained ViT models.
In this work, we aim to develop methods for training large-scale trackers with manageable resource requirements, thereby making more advanced large-scale models accessible to a broader research community and accelerating the evolution of visual tracking.

To achieve the goal, we draw inspiration from the advancements in Parameter-Efficient Fine-Tuning (PEFT) for large language models (LLMs), which are the first to encounter the prohibitive costs associated with full fine-tuning of large-scale models\cite{delta_tuning_overview, peft_survey}. PEFT methods fine-tune a relatively small number of parameters while keeping the rest frozen, significantly reducing computational and storage costs. While these methods have shown effectiveness in language models~\cite{llamaadapter} and vision-language models~\cite{t2i_adapter, tip_adapter}, their application to visual tracking, which will present unique challenges and potential domain gaps in our study, has been under-explored. This work is the first to investigate the application of PEFT in visual tracking, assessing its viability, and addressing domain-specific challenges. Among various PEFT research streams, we select Low-Rank Adaptation (LoRA)~\cite{lora}. LoRA achieves parameter-efficient fine-tuning by adding trainable rank decomposition matrices to certain dense layers of the frozen model, providing competitive performance compared to other PEFT approaches such as adapters~\cite{adapter, adapter_p, adapter_l} or prompt tuning~\cite{prompt_tuning, p_tuning, prefix_tuning}, without additional inference latency. 

However, unlike language models, the downstream task adaptation for vision models usually requires careful design upon the pre-trained backbone network~\cite{language_models_are_general_purpose_interfaces, gpt3}. Hence, we need a basic tracking framework as the starting point. We employ the one-stream framework~\cite{ostrack} as our baseline, because it brings minimal changes to the pre-trained ViT model, requiring only an image-pair capable positional encoding module and a head network. This architecture minimizes the extra parameters needed for training, aligning with the purpose of parameter-efficient fine-tuning. Additionally, the encoder-only Transformer architecture has been demonstrated to be more efficient in training trackers~\cite{swintrack}. 

Yet, it is \emph{not} as easy as our first intuition to apply LoRA to the one-stream tracking architecture, mainly due to two key issues below: \ding{172} \emph{Ineffective design of positional encoding module}. Most Transformer-based trackers adopt separate positional encodings for template and search region tokens. Albeit effective in full fine-tuning~\cite{ostrack}, such design is not well aligned with LoRA-style PEFT and thus results in suboptimal performance. We argue that this ineffectiveness comes from the disruption to the original structure of the pre-trained model, which is critical to preserve in PEFT, such as LoRA~\cite{lora}. \ding{173} \emph{Inductive biases}. We empirically identify the convolutional head in the one-stream tracker~\cite{ostrack} as a bottleneck in achieving convergence with LoRA. We argue that this is caused by strong inductive biases in convolutional network~\cite{vit}, making it hard to adapt the pre-trained model by fine-tuning a small subset of parameters using LoRA.

\noindent
\textbf{Our solution.} To overcome the above issues in employing LoRA to one-stream tracker~\cite{ostrack}, we present two embarrassingly \emph{simple but effective} designs. Specifically, in addressing issue \ding{172}, we draw inspiration from the well-known BERT~\cite{bert} and introduce the token type embedding in input embedding. Concretely, we assign specific types to tokens of template and search region. By doing so, we can directly leverage existing positional encoding modules of pre-trained ViT models without disrupting their original structure. Considering different resolutions of template and search regions, we further devise a multi-resolution absolute positional embedding adaptation strategy, enabling the support of multi-resolution inputs to pre-trained ViT models for better performance. To deal with issue \ding{173}, we propose a multilayer perceptron (MLP)-based anchor-free head network instead of the convolutional head network as in~\cite{ostrack}, for target classification and bounding box regression. Such a design effectively avoids the inductive biases when applying LoRA to our baseline one-stream tracker, enabling better performance with less computational overhead as shown in our experiments. 

It is worth noting that, despite the simplicity in implementation, our solution is specifically devised to employ LoRA better for improving visual tracking. Particularly, with our two designs, we propose LoRAT by applying LoRA to the one-stream OSTrack using various pre-trained ViTs, and extensively assess its performance on five large-scale benchmarks, including LaSOT~\cite{lasot}, LaSOT$_\mathrm{ext}$~\cite{lasot_ext}, TrackingNet~\cite{trackingnet}, GOT-10k~\cite{got10k}, and TNL2K~\cite{tnl2k}. Our best variant, LoRAT-g-378, sets a new record on LaSOT with a 0.762 SUC score when equipped with ViT-g~\cite{dinov2}, while the lightest variant, LoRAT-B-224, still achieves a comparable SUC score of 0.717 on LaSOT yet running at 209 FPS. In addition, the required training resources and time are manageable. The training time of different variants, from the light LoRAT-B-224 to strong LoRAT-g-378, ranges from 5.9 to 60 hours using 8 Nvidia V100 GPUs. Notably, our base LoRAT-B-224 can even be trained on a single customer-grade Nvidia RTX 4090 GPU within 11 hours with 605 inference FPS.

In summary, we make the following \textbf{contributions}: \ding{171} We, for the first time, propose leveraging LoRA to develop efficient and resource-friendly generic object tracking; \ding{170} We propose two simple yet effective designs to enable better adaption of LoRA for the tracking task; \ding{168} Our tracker LoRAT, by marrying LoRA to a one-stream tracker, achieves new state-of-the-art performance on multiple benchmarks with reasonable resource requirements.

\section{Related Work}
\label{sec:related_works}

\subsection{Paradigms of Transformer Tracker}
Visual tracking has witnessed significant progress in past decades~\cite{siamrpnPlus, dimp,prdimp, ocean, transt, stark, mixformer, swintrack, ostrack, seqtrack, artrack}, particularly with the integration of Transformer architectures.
Early exploration focuses on utilizing Transformer to replace or enhance components in traditional tracking frameworks. The work of~\cite{tmt} proposes the first Transformer tracker by using Transformer for feature fusion in Siamese~\cite{siamfc} and DCF~\cite{kcf, dimp} pipelines, respectively. 
TransT~\cite{transt} leverages self-attention for feature enhancement and cross-attention for feature fusion.
Subsequent research exploits the Transformer's modeling flexibility for improved template-search region relationship representation. The approaches of~\cite{tracker_inmo_module,sbt} explore feature interaction within the backbone network.
Stark~\cite{stark} and SwinTrack~\cite{swintrack} explore joint feature enhancement and feature interaction via Transformer but still separate backbone networks for feature extraction on template and search region. 

More recently, the one-stream tracking framework emerges as a significant evolution, with Mixformer~\cite{mixformer}, SimTrack~\cite{simtrack}, and OSTrack~\cite{ostrack} serving as the vanguard. This paradigm allows for earlier interaction between the template and search region, facilitating joint feature extraction and fusion. The self-attention mechanism of the Transformer architecture plays a crucial role in this context, enhancing the model's ability to capture complex dependencies.
Most subsequent Transformer trackers follow this paradigm.
GRM~\cite{tracker_grm} introduces adaptive token division to enable more flexible relation modeling.
DropTrack~\cite{droptrack} adaptively performs spatial-attention dropout in the frame reconstruction to facilitate temporal correspondence learning in videos. 
SeqTrack~\cite{seqtrack} and ARTrack~\cite{artrack} add a Transformer decoder on the top of the one-stream encoder to predict object bounding boxes in an auto-regressive fashion.

\subsection{Parameter Efficient Fine Tuning}
Parameter Efficient Fine Tuning (PEFT) is a solution to the challenge of fine-tuning large models, which becomes impractical due to high costs. PEFT enables efficient fine-tuning by adjusting a significantly smaller subset of parameters instead of the full model. PEFT methods primarily fall into two categories: prompt tuning-based~\cite{prefix_tuning, prompt_tuning, p_tuning, p_tuning_v2_acl} and adapter-based approaches~\cite{adapter, adapter_l, adapter_p}.
Prompt tuning-based approaches utilize continuous vectors as a part of input prompts, which are optimized through gradient descent during fine-tuning. Adapter-based approaches, on the other hand, introduce additional learnable modules either within or parallel to specific network components. One notable family of adapter-based methods is LoRA~\cite{lora} and its variants~\cite{qlora, adalora, lora+}. These methods apply low-rank matrices to approximate weight changes during fine-tuning and can be merged with pre-trained weights prior to inference, making them particularly efficient as they do not add any extra inference burden.

\section{Adapting ViT for LoRA-friendly Tracker}
\label{sec:methodology}

\begin{figure}[t]
    \centering
    \includegraphics[width=1.0\linewidth]{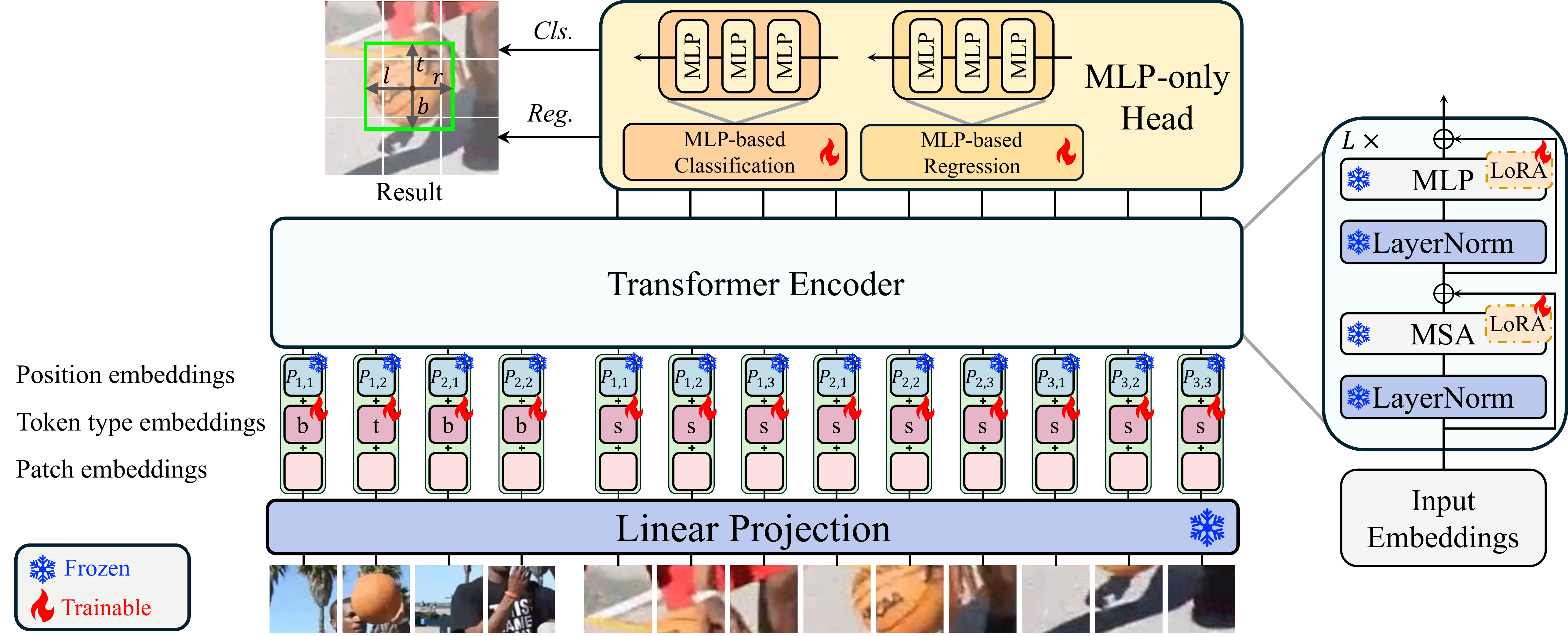}
    \caption{Architecture of LoRAT. The template and search region are first split and then projected as patch embeddings. Patch embeddings are added with shared position embeddings and token type embeddings as the input embeddings, which are then fed into Transformer encoder for joint feature extraction and fusion. The resulting representations are fed to the MLP-only head network for target classification and anchor-free-based bounding box regression. Most network components from the pre-trained ViT model are frozen during training, except for LoRA modules applied on the linear layers in the Transformer encoder, the token type embeddings, and the head network.}
    \label{fig:framework}
\end{figure}

This section meticulously outlines our approach, beginning with an explanation of our baseline, the one-stream tracking framework. We then introduce Low-Rank Adaptation (LoRA), a technique for efficient fine-tuning of large-scale models for specialized tasks, emphasizing its role in enhancing model adaptability with minimal computational overhead. Following this, we explore the architectural adjustments necessary to make our model compatible with LoRA, including modifications to the model's input embedding and an MLP-only head network.
An overview of our tracker is shown in Figure~\ref{fig:framework}.

\subsection{Preliminaries}

\subsubsection{One-Stream Tracker.} 
Our approach builds upon the one-stream tracking framework~\cite{ostrack}, which utilizes the Transformer's multi-head attention mechanism~\cite{transformer} to attend to different representation subspaces at various positions simultaneously, enabling joint feature extraction and fusion.

One-stream trackers follow the Siamese tracker framework~\cite{siamfc}, requiring a template image $T \in \mathbb{R}^{H_T \times W_T \times 3}$ and a search region image $S \in \mathbb{R}^{H_S \times W_S \times 3}$ as inputs.
The template image $T$ and the search region image $S$ are first projected by a learnable linear projection layer to obtain the template token embeddings and the search region token embeddings. Assuming the image patch resolution is $(P, P)$, the template image is divided into $N_T=N_T^H \times N_T^W$ image patches, and the search region image is divided into $N_S=N_S^H \times N_S^W$ image patches, where ${N_T^H, N_T^W, N_S^H, N_S^W}={(H_T, W_T, H_S, W_S)} / P$. The resulting patch embeddings are the template token embeddings $\mathbf{T} \in \mathbb{R}^{N_T \times d}$ and the search region embeddings $\mathbf{S} \in \mathbb{R}^{N_S \times d}$, where $d$ represents the hidden dimension of the network.

Subsequently, two positional embeddings, $\mathbf{E}_T$ and $\mathbf{E}_S$, are added to these token embeddings respectively, obtaining the template input embeddings $\mathbf{t}_0 \in \mathbb{R}^{N_T \times d}$ and the search region input embeddings $\mathbf{s}_0 \in \mathbb{R}^{N_S \times d}$:
\begin{align}
    \mathbf{t}_0 = \mathbf{T} + \mathbf{E}_T, ~~
    \mathbf{s}_0 = \mathbf{S} + \mathbf{E}_S.
\end{align}

Then, the template input embeddings $\mathbf{t}_0$ and the search region input embeddings $\mathbf{s}_0$ are concatenated as $\mathbf{z}_0$ and being fed into an $L$-layer Transformer encoder. The encoder comprises multi-head self-attention mechanisms and feed-forward neural networks. The output of the encoder, $\mathbf{z}_L \in \mathbb{R}^{(N_T + N_S) \times d}$, represents the joint feature representation of the template image and the search region image. Finally, a de-concatenation operation is performed to obtain the feature representations of the template image and the search region, denoted as $\mathcal{T}$ and $\mathcal{S}$, respectively. The search region feature representation $\mathcal{S}$ is then fed into the head network for further processing. The whole process is encapsulated by the following equations:
\begin{align}
    \mathbf{z}_0 &= \mathrm{Concat}(\mathbf{t}_0, \mathbf{s}_0), \\
    \mathbf{z^\prime}_\ell &= \op{MSA}(\op{LN}(\mathbf{z}_{\ell-1})) + \mathbf{z}_{\ell-1}, && \ell=1\ldots L, \\
    \mathbf{z}_\ell &= \op{MLP}(\op{LN}(\mathbf{z^\prime}_{\ell})) + \mathbf{z^\prime}_{\ell}, && \ell=1\ldots L, \\
    \mathcal{T}, \mathcal{S} &= \mathrm{DeConcat}(\mathbf{z}_L).
\end{align}

\subsubsection{Low-Rank Adaptation.} 
Low-Rank Adaptation (LoRA)~\cite{lora} is a technique for efficiently fine-tuning large-scale pre-trained models.
Typically, these models are initially trained on extensive datasets, providing broad foundational knowledge. To adapt the models to specific downstream tasks, models are fine-tuned with domain-specific data. Traditional fine-tuning is computationally intensive, as it involves a full forward and backward pass through the network, requiring significant computational resources and memory.

LoRA addresses these issues by proposing a low-rank matrix decomposition to represent weight updates, which can greatly reduce trainable weights and memory requirements. The principle behind LoRA is that while neural networks typically have full-rank weight matrices during pre-training, the updates required for fine-tuning to a specific task have a low ``intrinsic dimension''. Remarkably, in many applications, LoRA shows comparable performance to fully fine-tune the network.

In the context of fine-tuning weight updates (\(\Delta \Theta\)), LoRA employs considerably smaller matrices (\(\Delta \Phi\)), thus dramatically reducing the parameter space that needs to be learned. For a weight matrix update in the network \(\Delta \theta_i\), LoRA approximates it using:
\[
\Delta \theta_i \approx \Delta \phi_i = BA
\]
where \(B \in \mathbb{R}^{d \times r}\) and \(A \in \mathbb{R}^{r \times k}\), with the rank \(r \ll \min(d, k)\). This significantly reduces the learning complexity from learning \(d \times k\) parameters to learning only \((d + k) \times r\) parameters. 

Importantly, adopting LoRA does not increase the model's inference latency. After fine-tuning, the model's weights are updated by adding the low-rank approximations to the original weights. This approach simplifies the deployment of multiple task-specific models on top of a single large model because the updates (\(\Delta \Phi\)) are much smaller than the original weights (\(\Delta \Theta\)).

\subsection{LoRA-friendly Model Design}

\subsubsection{Decoupled Input Embedding.}
\label{subsec:input_emb}

ViT models require positional encodings to inject positional information~\cite{transformer}. Existing one-stream trackers utilize separate positional embeddings for template and search region tokens, which disrupt the structure of the pre-trained ViT model and lead to ineffectiveness in PEFT. To tackle this, we explore adaptations for multi-resolution positional embeddings and integrate token type embeddings.

\noindent
\emph{Token type embedding.}
We incorporate token type embedding (also known as segment embedding), originally proposed in the BERT\cite{bert} language model, into our tracker assigning unique embeddings for template and search region tokens.
This approach decouples token type identification from positional embeddings, improving parameter efficiency and flexibility.

Further, we extend the use of token type embedding to address specific challenges posed by the cropping strategy typically employed in Siamese trackers~\cite{siamfc}. The cropping strategy to generate tracking templates has several issues that may confuse the tracker: the aspect ratio of the target objects in the templates is variable; some target objects lack clear semantic meanings; some target objects in the templates can not be clearly distinguished from the background. Token type embedding can help alleviate these issues by explicitly annotating foreground and background parts within the template. We refer to this specific application of token type embedding as \emph{foreground object indication embedding}.

\noindent
\emph{Positional embedding.} 
Most ViT models employ the 1D absolute positional embedding~\cite{vit}, which only works for fixed resolutions. However, the one-stream trackers usually utilize an image pair of two different sizes as inputs. According to the PEFT-friendly model design principle, sharing the positional embedding across two images is essential. We explore strategies for reusing the standard 1D absolute positional embeddings effectively for different input sizes.
Both strategies regard the original 1D absolute embedding in a 2D view. Suppose the original 1D absolute positional embedding is $P$, the 2D view of $P$ can be $P_{2d}=[ p_{1,1}, ..., p_{1,w}; p_{2,1}, ..., p_{2,w}; p_{h,1}, ..., p_{h,w} ]$. 
For convenience, we assume that the resolution of the search region image is the native resolution of the ViT model, thus $P_{2d}$ can be directly applied to search region tokens, \ie the search region positional embedding $P_x = P_{2d}$. 

The first approach regards absolute positional embedding as continuous spatial position embedding. We use interpolation to resize $P_x$ to the desired resolution for the template, denoted as $P_z$. Thus, $P_z = \mathrm{interpolate}(P_x)$. We call this approach as \emph{interpolation-based} positional embedding adaptation. Notably, this technique is utilized by several Transformer-based trackers, such as OSTrack~\cite{ostrack}, to initialize distinct positional embeddings for template and search region tokens.

The second approach regards absolute positional embedding as discrete patch index embedding. Thus, with a smaller input resolution, the positional embedding for template tokens $P_z$ can be directly taken as a sub-matrix of the $P_x$. Specifically, the positional embedding for the template tokens is formed by selecting a block of entries corresponding to the size of the template image patches, $(N_T^H \times N_T^W)$, from the top left corner of the $P_x$. In this way, $P_z = [ p_{1,1}, ..., p_{1,N^W_T}; p_{2,1}, ..., p_{2,N^W_T}; p_{N^H_T,1}, ..., p_{N^H_T,N^W_T} ] $. We call this approach as \emph{slicing-based} positional embedding adaptation. 

In our ablation experiments, we found the \emph{slicing-based} adaptation of absolute positional embeddings outperformed the \emph{interpolation-based} approach. Consequently, we adopted the \emph{slicing-based} method as our primary strategy for adapting absolute positional embeddings within our tracker.

For other widely utilized positional encoding schemes, such as sinusoidal positional encoding~\cite{vit}, relative positional embedding\cite{shaw2018relpos}, or Rotary Positional Embedding\cite{roformer}, these do not inherently target specific resolutions and can be directly applied to template tokens and search region tokens.

The final input embeddings are constructed by adding the positional embedding and token type embedding to the patch embeddings:
\begin{eqnarray}
\mathbf{E}_T^{(i,j)} &=& \left\{ \begin{array}{ll}
\mathbf{E}_{pos}^{(i,j)} + \mathbf{E}^{T_{o}}_{type}, & \textrm{\;\;\;if }\ \mathbf{T}^{(i,j)}\textrm{ 
belongs target,}\\
\mathbf{E}_{pos}^{(i,j)} + \mathbf{E}^{T_{b}}_{type}, & \;\;\; \textrm{otherwise} 
\end{array} \right.  \\
\mathbf{E}_S^{(i,j)} &=& \mathbf{E}_{pos}^{(i,j)} + \mathbf{E}^{S}_{type},
\end{eqnarray}
where $\mathbf{E}^{T_{o}}_{type} \in \mathbb{R}^d$, $\mathbf{E}^{T_{b}}_{type} \in \mathbb{R}^d$, and $\mathbf{E}^{S}_{type} \in \mathbb{R}^d$ are the token type embedding of the template foreground (target object) tokens, the template background tokens, and search region tokens respectively. \textbf{Notice}, the positional encoding $\mathbf{E}_{pos}^{(i,j)}$ is \textbf{shared} between template and search region.

\subsubsection{MLP-only Head Network.}
\noindent
To mitigate potential inductive biases inherent in convolutional heads, such as locality assumptions on data structure~\cite{vit}, which may impede convergence during LoRA-based fine-tuning, we utilize a multi-layer perceptron (MLP) only head.

The head network generates a redundant set of bounding boxes, each accompanied by a classification score. The classification scores are used to rank the bounding boxes. The bounding box with the highest classification score is considered as the final output. To achieve this, the head network is divided into two branches: one for classification and another for bounding box regression, each consisting of a three-layer MLP. Both branches process the feature map of the search area $\mathcal{S}$, outputted from the Transformer encoder. The head network employs a center-based, anchor-free style~\cite{fcos} to accelerate the training coverage and boost the tracking performance.

\section{Experiments}
\label{sec:exp}

\subsection{Experimental Setup}

LoRAT is trained and evaluated on 8 $\times$ NVIDIA V100 GPUs. This section reports the main experimental setup, with more details in \textbf{supplementary material}.

\noindent
{\bf Model.} We present six variants of LoRAT, namely B-224, B-378, L-224, L-378, g-224, and g-378. These variants are based on three sizes of the ViT backbone networks: ViT-B~\cite{vit}, ViT-L~\cite{vit}, and ViT-g~\cite{dinov2}. Each size is configured with two different input resolutions. For the -224 variants, the template size is set to $[112 \times 112]$, and the search region size is set to $[224 \times 224]$; For the -378 variants, the template size is set to $[196 \times 196]$, the search region size is set to $[378 \times 378]$. By default, all variants employ the DINOv2\cite{dinov2} pre-trained weights.

\noindent
{\bf Training.} The trackers are trained using the training splits of LaSOT\cite{lasot}, TrackingNet~\cite{trackingnet}, GOT-10k~\cite{got10k} (1k sequences removed for fair comparison following~\cite{stark,ostrack}, as they overlap with videos used in \cite{VOT_TPAMI}), and COCO 2017~\cite{coco}.
For the GOT-10k benchmark, trackers are trained using only the GOT-10k training split, following the protocol in \cite{got10k}.
The trackers are trained for 170 epochs, each epoch containing 131,072 image pairs. For the GOT-10k benchmark, the training epochs are reduced to 100 to mitigate over-fitting.
Each GPU holds 16 samples for an iteration, resulting in a batch size of 128.
Following the suggestion in \cite{qlora}, LoRA is applied on all linear layers of the ViT backbone, including four projection matrices in the multi-head self-attention module and two projection matrices in the MLP module. The rank $r$ of LoRA is set to 64 for all variants.
All LoRA linear layers are initialized using the same method as the BERT~\cite{bert} linear layer, with a truncated normal distribution with std 0.02.

\noindent
{\bf Inference.} LoRAT follows conventional steps of Siamese tracking~\cite{siamfc} during inference: the template is cropped from the first frame of the video sequence, centering around the target object; the search area is cropped from the current tracking frame, with the center determined by the bounding box of target object predicted from the previous frame. A Hanning window is added to the classification response map $\mathbf{R}$ to suppress large displacements.

\begin{table*}[t] 
\caption{Benchmarking our tracker on five large-scale challenging datasets. For GOT-10k evaluation, all the methods follow the one-shot protocol, training only on the train split of GOT-10k. Bold indicates the best results.}
\label{tbl:overall_results}

\begin{center}
    
\resizebox{1.0\textwidth}{!}{
\begin{tabular}{ r | ccc | ccc | ccc | ccc | cc }
\toprule
 \multirow{2}{*}{\begin{tabular}[r]{@{}c@{}}Tracker\end{tabular}} & \multicolumn{3}{c|}{LaSOT \cite{lasot}} & \multicolumn{3}{c|}{LaSOT$_{\text{ext}}$ \cite{lasot_ext}} & \multicolumn{3}{c|}{TrackingNet \cite{trackingnet}} & \multicolumn{3}{c|}{GOT-10k \cite{got10k}} & \multicolumn{2}{c}{TNL2K \cite{tnl2k}} \\
                                & 
{\small SUC}        & {\small P$_\mathrm{Norm}$}     & {\small P}              & {\small SUC}            & {\small P$_\mathrm{Norm}$}     & {\small P}              &
{\small SUC}            & {\small P$_\mathrm{Norm}$}     & {\small P}            & {\small AO}           & {\small SR$_\mathrm{0.5}$}     & {\small SR$_\mathrm{0.75}$}    &
{\small SUC}            & {\small P}              \\
\midrule
TransT \cite{transt}            &
64.9           & 73.8           & 69.0           & -              & -              & -              &
81.4           & 86.7           & 80.3           & 67.1           & 76.8           & 60.9           &
50.7           & 51.7           \\
AutoMatch \cite{automatch}      &
58.3           & -              & 59.9           & 37.6           & -              & 43.0           &
76.0           & -              & 72.6           & 65.2           & 76.6           & 54.3           &
47.2           & 43.5           \\
STARK \cite{stark}              &
67.1           & 77.0           & -              & -              & -              & -              &
82.0           & 86.9           & -              & 68.8           & 78.1           & 64.1           &
-              & -              \\
KeepTrack \cite{keeptrack}      &
67.1           & 77.2           & 70.2           & 48.2           & -              & -              &
-              & -              & -              & -              & -              & -              &
-              & -              \\
MixFormer \cite{mixformer}	&
70.1           & 79.9           & 76.3           & -              & -              & -              &
83.9           & 88.9           & 83.1           & -              & -              & -              &
-              & -              \\
SBT \cite{sbt}			        &
66.7           & -              & 71.1           & -              & -              & -              &
-              & -              & -              & 70.4           & 80.8           & 64.7           &
-              & -              \\
AiATrack \cite{aiatrack}        &
69.0           & 79.4           & 73.8           & 47.7           & 55.6           & 55.4           &
-              & -              & -              & 69.6           & 80.0           & 63.2           &
-              & -              \\
SimTrack \cite{simtrack}        &
70.5           & 79.7           & -              & -              & -              & -              &
-              & -              & -              & 69.8           & 78.8           & 66.0           &
55.6           & 55.7           \\
OSTrack  \cite{ostrack}         &
71.1           & 81.1           & 77.6           & 50.5           & 61.3           & 57.6           &
83.9           & 88.5           & 83.2           & 73.7           & 83.2           & 70.8           &
55.9           & 56.7           \\
SwinTrack     \cite{swintrack}  &
71.3           & -              & 76.5           & 49.1           & -              & 55.6           &
84.0           & -              & 82.8           & 72.4           & 80.5           & 67.8           &
55.9           & 57.1           \\
DropTrack \cite{droptrack}      &
71.8           & 81.8           & 78.1           & 52.7           & 63.9           & 60.2           &
-              & -              & -              & 75.9           & 86.8           & 72.0           &
56.9           & 57.9           \\
SeqTrack      \cite{seqtrack}   &
72.5           & 81.5           & 79.3           & 50.7           & 61.6           & 57.5           &
85.5           & 89.8           & 85.8           & 74.8           & 81.9           & 72.2           &
57.8           & -              \\
ARTrack      \cite{artrack}     &
73.1           & 82.2           & 80.3           & 52.8           & 62.9           & 59.7           &
85.6           & 89.6           & 86.0           & 78.5           & 87.4           & 77.8           &
60.3           & -              \\
CiteTracker \cite{citetracker}  &
69.7           & 78.6           & 75.7           & -              & -              & -              &
84.5           & 89.0           & 84.2           & 74.7           & 84.3           & 73.0           &
57.7           & 59.6           \\
ROMTrack~\cite{ROMTrack}        &
71.4           & 81.4           & 78.2           & 51.3           & 62.4           & 58.6           &
84.1           & 89.0           & 83.7           & 74.2           & 84.3           & 72.4           &
-              & -              \\
MixViT~\cite{mixformer_extend}&
72.4           & 82.2           & 80.1           & -              & -              & -              &
85.4           & \textbf{90.2}           & 85.7           & 75.7           & 85.3           & 75.1           &
-              & -              \\
\midrule
LoRAT-B-224                 &
71.7           & 80.9           & 77.3           & 50.3           & 61.6           & 57.1           &
83.5           & 87.9           & 82.1           & 72.1           & 81.8           & 70.7           &
58.8           & 61.3           \\ 
LoRAT-B-378                 &
72.9           & 81.9           & 79.1           & 53.1           & 64.8           & 60.6           &
84.2           & 88.4           & 83.0           & 73.7           & 82.6           & 72.9           &
59.9           & 63.7           \\
LoRAT-L-224                 &
74.2           & 83.6           & 80.9           & 52.8           & 64.7           & 60.0           &
85.0           & 89.5           & 84.4           & 75.7           & 84.9           & 75.0           &
61.1           & 65.1           \\
LoRAT-L-378                 &
75.1           & 84.1           & 82.0           & \textbf{56.6}           & \textbf{69.0}           & \textbf{65.1}           &
85.6           & 89.7           & 85.4           & 77.5           & 86.2           & 78.1           &
62.3           & 67.0           \\
LoRAT-g-224                 &
74.9           & 84.5           & 82.3           & 53.3           & 65.4           & 61.1           &
85.2           & 89.8           & 85.1           & 77.7           & 87.7           & 77.7           &
61.8           & 66.6           \\
LoRAT-g-378                 &
\textbf{76.2}           & \textbf{85.3}           & \textbf{83.5}           & 56.5           & \textbf{69.0}           & 64.9           &
\textbf{86.0}           & \textbf{90.2}           & \textbf{86.1}           & \textbf{78.9}           & \textbf{87.8}           & \textbf{80.7}           &
\textbf{62.7}           & \textbf{67.8}           \\
\bottomrule
\end{tabular} }
\end{center}
\end{table*}

\begin{SCtable}
\caption{Comparison on efficiency with state-of-the-art Transformer trackers. The values in parentheses for the \# params of our trackers represent LoRA and extra components (token type embedding and head), respectively. The speed of all trackers was re-evaluated on our machine.}
\label{tbl:efficiency_table}
\centering
\resizebox{0.65\textwidth}{!}{
\begin{tabular}{ r | c | c | c }
\toprule
Tracker                          & Speed ({\it fps}) & MACs (G)  & \#Params (M) \\
\midrule
SwinTrack-B-384~\cite{swintrack} & 45          & 69.7      & 91           \\
OSTrack-256~\cite{ostrack}       & 130         & 21.5      & -            \\
OSTrack-384~\cite{ostrack}       & 68          & 48.3      & -            \\
SeqTrack-B256~\cite{seqtrack}    & 38          & 66        & 89           \\
SeqTrack-L384~\cite{seqtrack}    & 6           & 524       & 309          \\
\midrule
LoRAT-B-224                      & 209         & 30        & 99 (11, 2)   \\
LoRAT-B-378                      & 151         & 97        & 99 (11, 2)   \\
LoRAT-L-224                      & 119         & 103       & 336 (28, 4)  \\
LoRAT-L-378                      & 63          & 325       & 336 (28, 4)  \\
LoRAT-g-224                      & 50          & 378       & 1216 (71, 9) \\
LoRAT-g-378                      & 20          & 1161      & 1216 (71, 9) \\
\bottomrule
\end{tabular} }
\end{SCtable}

\subsection{Quantitative Comparison with the State-of-the-Art Models}
This section comprehensively evaluates the proposed LoRAT, benchmarking it against the latest Transformer-based tracking models across five datasets. The official evaluation protocols are followed to ensure a fair and accurate comparison. The results are detailed in Table~\ref{tbl:overall_results}. Additionally, an efficiency comparison with several state-of-the-art Transformer trackers is presented in Table~\ref{tbl:efficiency_table}.

\noindent
{\bf LaSOT}~\cite{lasot} is a large scale benchmark containing 280 test videos. 
From Table~\ref{tbl:overall_results}, our smallest variant, LoRAT-B-224, achieves a competitive SUC score of 0.717. LoRAT-L-224 surpasses all other state-of-the-art models with an SUC score of 0.742. Our largest variant, LoRAT-g-384, utilizing the most substantial backbone network ever applied in visual tracking, sets a new benchmark with an SUC score of 0.762, significantly outperforming the previous best, ARTrack~\cite{artrack}, by 3.1\%.

\noindent
{\bf LaSOT${_\mathrm{ext}}$}~\cite{lasot_ext} is an extension of LaSOT with 150 additional highly challenging sequences.
According to Table~\ref{tbl:overall_results}, LoRAT-B-378, using the ViT-B backbone, achieves a SUC score of 0.531, surpassing the former leader, ARTrack~\cite{artrack}, by 0.4\%. LoRAT-L-378, with the ViT-L backbone, reaches a new high with an SUC score of 0.566. Our largest variant, LoRAT-g-378, achieves a comparable SUC score of 0.565.

\noindent
{\bf TrackingNet}~\cite{trackingnet} is a large-scale object tracking benchmark. The test split of TrackingNet includes 511 video sequences. The results in Table~\ref{tbl:overall_results} show that all variants of our tracker achieve competitive SUC scores, ranging from 0.835 to 0.860. Notably, the largest variant, LoRAT-g-378, records the highest SUC score of 0.860, outperforming the previous top-performing model by 0.4\%.

\noindent
{\bf GOT-10k}~\cite{got10k} provides 180 videos for testing, with a strict protocol that only allows the GOT-10k training split for model training. As reported in Table~\ref{tbl:overall_results}, the size of the model significantly influences performance on this benchmark, much more than the input resolution. our largest variant, LoRAT-g-378, achieves a new state-of-the-art AO score of 78.9\%.

\noindent
{\bf TNL2k}~\cite{tnl2k} is a recent tracking dataset with 700 test videos. As Table~\ref{tbl:overall_results} indicates, LoRAT-L-378 achieves a 2.0\% gain over ARTrack~\cite{artrack} with a 0.623 SUC score. LoRAT-g-378 achieves the best performance with an SUC score of 0.627.

\noindent
{\bf Efficiency Comparison.}
In Table~\ref{tbl:efficiency_table}, we compare the efficiency of LoRAT against other Transformer-based trackers across key metrics, including running speed, floating point operations, and number of parameters. LoRAT benefits from a highly efficient one-stream architecture and the zero inference cost of LoRA, resulting in an impressive running speed compared to its floating point operations. Our most compact variant, LoRAT-B-224, achieves state-of-the-art (SOTA) performance while operating at 209 frames per second (fps). The intermediate LoRAT-L-378 variant surpasses most competitors, maintaining a speed of 63 fps. Notably, the LoRAT-g-378, equipped with the largest publicly available ViT backbone, still operates at a practical speed of 20 fps.
Besides, our tracker also exhibits efficiency in storage requirements. The trainable components of the LoRAT-g-378 variant (which can be considered as a visual tracking adapter for ViT), including LoRA, token type embedding, and the head network, require only 80 million parameters. This represents a significant reduction from the 1.1 billion parameters of the full ViT-g model~\cite{dinov2}.

\subsection{Ablation Experiments}
We conduct detailed ablation studies to dissect the contributions of various components within LoRAT. These models in the experiments are evaluated on three benchmarks: LaSOT~\cite{lasot}, LaSOT$_\mathrm{ext}$~\cite{lasot_ext}, and TNL2k~\cite{tnl2k}.

\begin{table}[!t]
\caption{Ablation on the performance improvement of LoRA-based fine-tuning on our tracker. All variants in the table are trained with full fine-tuning. The results are compared with our default settings, \ie the variants trained with LoRA. }
\label{tbl:lora_vs_ft}
\centering
\scalebox{1.0}{
\begin{tabular}{ c | cc | cc | cc }
\toprule
\multirow{2}{*}{\begin{tabular}[r]{@{}c@{}}Variant\end{tabular}} & 
\multicolumn{2}{c|}{LaSOT \cite{lasot}} & \multicolumn{2}{c|}{LaSOT$_{\text{ext}}$ \cite{lasot_ext}} & \multicolumn{2}{c}{TNL2K \cite{tnl2k}}
\\
                                & 
{\small SUC}        & {\small P}              & {\small SUC}            & {\small P}              &
{\small SUC}        & {\small P}     \\
\midrule
B-224             &
70.9 (\textbf{$\downarrow$0.8}) & 76.2 (\textbf{$\downarrow$1.1}) & 50.0 (\textbf{$\downarrow$0.3}) & 57.1 (\textbf{$\downarrow$0.0})    &
58.1 (\textbf{$\downarrow$0.7}) & 60.1 (\textbf{$\downarrow$1.2})    \\
B-378             &
73.2 (\textbf{$\uparrow$0.3})    & 78.8 (\textbf{$\downarrow$0.3})    & 52.9 (\textbf{$\downarrow$0.2})    & 60.5 (\textbf{$\downarrow$0.1})    &
60.0 (\textbf{$\uparrow$0.1})    & 63.3 (\textbf{$\downarrow$0.4})    \\ 
L-224             &
73.0 (\textbf{$\downarrow$1.2})    & 79.3 (\textbf{$\downarrow$1.6})    & 52.6 (\textbf{$\downarrow$0.2})    & 59.8 (\textbf{$\downarrow$0.2})    &
60.8 (\textbf{$\downarrow$0.3})    & 64.3 (\textbf{$\downarrow$0.8})    \\ 
L-378             &
74.9 (\textbf{$\downarrow$0.2})    & 81.8 (\textbf{$\downarrow$0.2})    & 55.2 (\textbf{$\downarrow$1.4})    & 63.1 (\textbf{$\downarrow$2.0})    &
61.7 (\textbf{$\downarrow$0.6})    & 65.7 (\textbf{$\downarrow$1.3})    \\ 
\bottomrule
\end{tabular} }
\end{table}

\noindent
{\bf Performance improvement of LoRA.}
We report the performance of LoRAT with full fine-tuning setting in Table~\ref{tbl:lora_vs_ft}, with a quantified comparison to LoRA-based fine-tuning. As shown in Table~\ref{tbl:lora_vs_ft}, the results demonstrate that LoRA significantly enhances the tracker’s performance across all variants. This performance improvement indicates the effectiveness of our design in leveraging parameter-efficient fine-tuning strategies. By integrating LoRA, we successfully mitigate catastrophic forgetting~\cite{catastrophic_forgetting}, which is a significant problem when adapting pre-trained models to downstream tasks~\cite{delta_tuning_overview, controlnet} like visual tracking. 

\noindent
{\bf Input embedding.} 
This ablation verifies the effectiveness of the proposed \emph{decoupled} input embedding scheme. The experiments are conducted on the ViT-L~\cite{vit} backbone with two input resolution settings.
From the results in Table~\ref{tbl:input_emb}, we observe that:
1) Freezing positional embedding shows slightly better performance than the unfreezing one (\ding{176} \vs \  \ding{172}). 2) Adopting ``shared positional embedding'', ``token type embedding'' and ``foreground indication embedding'' (our model) respectively outperforms the ``separated positional embedding'', ``non-token embedding'' and ``non-foreground indication embedding'' version (\ding{176} - \ding{179}). 3) Unfreezing the position embedding will degrade the effectiveness of our proposed foreground indicating embedding (\ding{175} \vs \ \ding{174}, \ding{179} \vs \ \ding{178}) 4) With a higher resolution of input images, foreground indication embedding significantly boost the performance of our tracker (\ding{181} \vs \  \ding{180}).

\begin{table}[!t]

\caption{Ablation on the input embedding configurations. All experiments are conducted on ViT-L~\cite{vit} backbone with two input resolution settings and with LoRA-based fine-tuning. We examine five factors:
1) Freezing positional embeddings during training;
2) Sharing positional embeddings between template and search region through \emph{slicing-based} positional embedding adaptation, or two separate positional embeddings are initialized with \emph{interpolation-based} positional embedding adaptation;
3) Using token type embeddings;
4) Incorporating foreground object indication embeddings;
5) Input resolution (224 \vs 378). }

\label{tbl:input_emb}

\centering
\footnotesize
\scalebox{1.0}{
\begin{tabular}{ c | c | c | c | c | c | cc | cc | cc }
\toprule
\multirow{2}{*}{} & \multirow{2}{*}{\shortstack{Frozen \\ P. Emb.}} & \multirow{2}*{\shortstack{ Shared \\ P. Emb.}} & \multirow{2}*{\shortstack{Type \\ Emb.}} & \multirow{2}*{\shortstack{Foreg. \\ Indic.}} & \multirow{2}{*}{Res.} & \multicolumn{2}{c|}{LaSOT~\cite{lasot}} & \multicolumn{2}{c|}{LaSOT$_\mathrm{ext}$~\cite{lasot_ext}} & \multicolumn{2}{c}{TNL2K~\cite{tnl2k}}  \\
& &    &  & &  & SUC & P & SUC & P & SUC & P   \\
\midrule
 \ding{172} & &  & &                                           & 224 & 73.9  & 80.4 & 51.9 & 59.0 & 60.6 & 64.3 \\
 \ding{173} & & \checkmark & &                                 & 224 & 73.9  & 80.6 & 52.7 & 59.8 & 60.8 & 64.8 \\
 \ding{174} & & \checkmark & \checkmark &                      & 224 & 74.2  & 81.1 & 52.9 & 60.2 & 60.8 & 64.7 \\
 \ding{175} & & \checkmark & \checkmark & \checkmark           & 224 & 74.0 & 80.8 & 52.8 & 60.1 & 61.2 & 65.5 \\
 \ding{176} &\checkmark &  & &                                 & 224 & 73.5 & 80.2 & 51.8 & 59.1 & 60.7 & 64.4 \\
 \ding{177} & \checkmark & \checkmark & &                      & 224 & 73.8  & 80.6 & 53.7 & 61.4 & 60.6 & 64.5  \\
 \ding{178} &\checkmark & \checkmark & \checkmark &            & 224 & 74.0 & 80.7 & 52.4 & 59.6 & 60.7 & 64.7  \\
 \ding{179} &\checkmark & \checkmark & \checkmark & \checkmark & 224 & 74.2 & 80.9 & 52.8 & 60.0 & 61.1 & 65.1 \\
 \ding{180} &\checkmark & \checkmark & \checkmark & & 378 &
 74.4 & 82.6 & 55.2 & 63.2 & 62.3 & 67.1 \\
 \ding{181} & \checkmark & \checkmark & \checkmark & \checkmark & 378 &
 75.1 & 82.0 & 56.6 & 65.1 & 62.3 & 67.0 \\
\bottomrule
\end{tabular}}
\end{table}

\noindent
{\bf MLP-only head network.} We evaluate the performance of MLP-only and convolutional head networks under LoRA-based and full fine-tuning settings. As shown in Table~\ref{tbl:abl_head}, the OSTrack~\cite{ostrack} convolutional head yields similar competitive results to the MLP-only head in the full fine-tuning setting but fails to converge in the LoRA-based fine-tuning setting.  We attribute this to the inductive biases in the convolution layer, which may not satisfy the low "intrinsic dimension" assumption required by low-rank adaptation. Additionally, an MLP-only head offers better computational efficiency.

\begin{table*}[!t]\footnotesize
\caption{Ablation on head modules. Conv head: Replace the 3-layers MLP-base head in our tracker with 5 stacked Conv-BN-ReLU layers; OSTrack~\cite{ostrack} head: Replace the head in our tracker with OSTrack~\cite{ostrack} convolution-based head. }
\label{tbl:abl_head}
\centering
\scalebox{1.}{
\begin{tabular}{ c | ccc | ccc | cc | c | c }
\toprule
\multirow{2}{*}{Variant} & 
  \multicolumn{3}{c|}{LaSOT~\cite{lasot}} &
  \multicolumn{3}{c|}{LaSOT$_{\text{ext}}$~\cite{lasot_ext}} &
  \multicolumn{2}{c|}{TNL2k~\cite{tnl2k}} &
 Speed & 
 \#Params\\  
& 
{\small SUC}        & {\small P$_\mathrm{Norm}$} & {\small P}          & {\small SUC}        & {\small P$_\mathrm{Norm}$} & {\small P}          &
{\small SUC}        & {\small P}   & \emph{fps} &  M     \\ 
\midrule
\multicolumn{2}{@{}l}{\emph{LoRA:}} \\
\midrule
\multicolumn{1}{l|}{L-224} & 
74.2           & 83.6           & 80.9           & 52.8           & 64.7           & 60.0  & 61.1 & 65.1 & 119 & 336 \\
\multicolumn{1}{l|}{+ Conv head}  &
4.7 & 6.4 & 1.6 & 4.5 & 9.4 & 0.8 & 0.0 & 0.1 & 119 & 336 \\
\multicolumn{1}{l|}{+ OSTrack~\cite{ostrack} head}   &
5.4 & 5.1 & 1.3 & 4.9 & 9.0 & 1.2 & 0.1 & 0.1 & 119 & 340 \\
\midrule
\multicolumn{2}{@{}l}{\emph{Full Fine-tune:}} \\
\midrule
\multicolumn{1}{l|}{L-224} & 
73.0 & 81.8 & 79.3 & 52.6 & 64.1 & 59.8 & 60.8 & 64.3 & 119 & 307 \\ 
\multicolumn{1}{l|}{+ Conv head} &
71.7 & 80.5 & 78.0 & 51.7 & 63.1 & 59.0 & 58.8 & 62.0 & 109 & 309 \\
\multicolumn{1}{l|}{+ OSTrack~\cite{ostrack} head} &
72.4 & 80.9 & 78.5 & 51.9 & 63.2 & 59.3 & 59.3 & 62.6 & 104 & 311 \\ 
\bottomrule
\end{tabular} }
\end{table*}

\noindent
\begin{minipage}[t]{.42\textwidth}
\centering
\captionof{table}{OSTrack variants implemented using official codes.}\label{tbl:ostrack_aligned}
\resizebox{0.82\textwidth}{!}{
\begin{tabular}{ c | c | c }
\toprule
{Variant} & 
{Pre-training} & 
{LaSOT {\small SUC}}
\\
\midrule
B-256 & MAE &
68.7
\\
L-256 & MAE &
70.3
\\
B-224 & DINOv2 &
66.2 
\\
L-224 & DINOv2 &
67.1
\\
\bottomrule
\end{tabular} }
\captionof{table}{Training time and max GPU memory usage of our tracker.} 
\label{tbl:lora_vs_ft_resource}
\resizebox{0.86\textwidth}{!}{
\begin{tabular}{ c | c | c | c }
\toprule
&
Variant & 
{Time(h)} &
{Memory(GB)}
\\
\midrule
\multirow{6}{*}{\rotatebox[origin=c]{90}{LoRA}}
& B-224                           &
5.9 & 2.4
\\
& B-378                           &
12.2 & 5.7
\\
& L-224                           &
10.8 & 6.6
\\
& L-378                           & 
32.2 & 14.9
\\
& g-224                           &
22.3 & 14.0
\\
& g-378                           &
60.0 & 25.8
\\
\midrule
\multirow{6}{*}{\rotatebox[origin=c]{90}{\tiny Full Fine-tuning}}
& B-224            &
5.9 & 3.2
\\
& B-378            &
12.5 & 6.5
\\
& L-224             &
12.1 & 10.0
\\
& L-378             &
34.1 & 19.1
\\
& g-224             &
29.3 & 27.1
\\
& g-378             &
\multicolumn{2}{c}{Out of Memory}
\\
\bottomrule
\end{tabular} }
\end{minipage}%
\hspace{0.02\textwidth}
\begin{minipage}[t]{.53\textwidth}
\centering  
\captionof{table}{Ablation on the impact of ViT backbone with various pre-training methods. The base is B-224 trained with \emph{full fine-tuning}.}\label{tbl:vit_pretrain}
\resizebox{1.0\textwidth}{!}{
\begin{tabular}{ r | cc | cc | cc }
\toprule
 \multirow{2}{*}{Pre-training} &
 \multicolumn{2}{c|}{LaSOT~\cite{lasot}} & \multicolumn{2}{c|}{LaSOT$_{\text{ext}}$~\cite{lasot_ext}} &
 \multicolumn{2}{c}{TNL2K~\cite{tnl2k}} \\
                                &
{\small SUC} & {\small P} & {\small SUC} & {\small P}  &
{\small SUC}        & {\small P}          \\
\midrule
MAE \cite{mae}                  &
70.3 & 75.3 & 49.2 & 56.0 & 56.9 & 58.6           \\ 
CLIP \cite{clip}                &
70.1           & 75.0           & 48.7           & 55.0           &
56.8           & 58.3              \\
EVA-02 \cite{eva02}             &
70.7 & 76.2 & 49.8 & 56.9 & 56.8 & 58.6           \\
DINOv2 \cite{dinov2}            &
70.9           & 76.2           & 50.0           & 57.1           &
58.1           & 60.1           \\
\bottomrule
\end{tabular} }
\captionof{table}{Training time, max training GPU memory usage, max training batch size and inference FPS of our tracker on a single customer-grade NVIDIA RTX 4090 GPU.}\label{tbl:resource_4090}
\resizebox{1.\textwidth}{!}{
\begin{tabular}{ c | c | c | c | c | c }
\toprule
 &
{Variant} & 
{Time(h)} &
{Memory(GB)} &
{Batch} &
{Speed (\emph{fps})}
\\
\midrule
\multirow{4}{*}{\rotatebox[origin=c]{90}{LoRA}}
& B-224                              &
11.0  & 14.1 & 128   & 605
\\
& B-378                              &
33.8  & 20.3 & 64  & 285
\\
& L-224                              &
34.0  & 19.5 & 64  & 289
\\
& L-378                              &
110.3 & 14.7 & 16 & 150
\\
\midrule
\multirow{4}{*}{\rotatebox[origin=c]{90}{\tiny Full Fine-tuning}}
& B-224              &
11.7  & 16.1 & 128 & 605
\\
& B-378              &
36.5  & 12  & 32 & 285
\\
& L-224              &
38.5  & 13.7 & 32 & 289
\\
& L-378              &
117.8 & 18 & 16 & 150
\\
\bottomrule
\end{tabular} }
\end{minipage}
\bigskip

\noindent
{\bf Aligned comparison with OSTrack.} We provide an aligned performance comparison with OSTrack~\cite{ostrack} by replacing only the backbone of OSTrack. For a fair comparison, OSTrack variants are trained without the early candidate elimination module. Input resolutions of variants are adjusted to align with the patch size of the ViT backbone. Results in Table~\ref{tbl:ostrack_aligned} indicate that all OSTrack variants underperform compared to our model. Besides, OSTrack models using the DINOv2 backbone demonstrate weaker performance compared to their default MAE backbone configuration. These results demonstrate the effectiveness of our model design.

\noindent
{\bf Different Pre-training Methods.} We evaluate the performance of four ViT variants, each pre-trained using different methodologies: MAE~\cite{mae}, CLIP~\cite{clip}, EVA02~\cite{eva02}, and DINOv2~\cite{dinov2}, when serving as the backbone of our tracker. According to Table~\ref{tbl:vit_pretrain}, the self-supervised DINOv2 pre-training model consistently outperforms other models on all tested datasets. Besides, this superiority, as reflected against the outcomes from Table~\ref{tbl:ostrack_aligned}, illustrates that our model design effectively unleashes the power of large-scale pre-training.

\noindent
{\bf Training Efficiency of LoRA.} The efficiency of LoRA's fine-tuning process is evaluated on two different hardware setups: one with 8 NVIDIA V100 GPUs and another with a single NVIDIA 4090 GPU, as shown in Tables~\ref{tbl:lora_vs_ft_resource} and \ref{tbl:resource_4090}, respectively. The results reveal that fine-tuning with LoRA significantly reduces both training time and GPU memory usage, particularly for larger models.


\section{Conclusion}
\label{sec:conclusion}

In this work, we introduce a novel approach to visual tracking by integrating Low Rank Adaptation (LoRA) within the one-stream tracking framework. Our contributions include two innovative designs that enable effective adaptation of pre-trained models for tracking, significantly reducing the resource requirements. The proposed tracker, LoRAT, not only achieves state-of-the-art performance on multiple benchmarks but also demonstrates the feasibility of training advanced tracking models with manageable resources. This study bridges the gap between advanced large-scale models and resource accessibility, marking a pivotal step towards sustainable advancements in the field of visual tracking. \\

\noindent\textbf{Acknowledgments.} This work is supported by Pengcheng Laboratory Research Project No. PCL2023A08 and No. U20B2052. Heng Fan, Zhipeng Zhang and Haibin Ling receive no financial support for the research, authorship, and/or publication of this article.

%
%
\bibliographystyle{splncs04}
\bibliography{main}

\clearpage

\section*{Appendix}
\appendix
\label{sec:appendix}
\section{More Implementation Details}

This section provides additional details on the implementation of our proposed LoRAT tracker.
The hyper-parameters used during training and testing are summarized in Table~\ref{tbl:hyperparameters}.

\subsection{General Settings}

\noindent
{\bf Input Data.}
Following common practice in Siamese Trackers~\cite{siamfc}, the template and search region are cropped from input images.
The template has a background area factor of 2, while the search region has a background area factor of 4 for -224 variants and 5 for -378 variants, as in \cite{ostrack}.

\noindent
{\bf Training.}
We apply common data augmentation strategies to the image pairs, including random color jittering and horizontal flipping. Additionally, we employ the 3-Augment~\cite{deit_3} strategy, with a minor modification ensuring that the images in a pair are applied with the same transformation.
The models are trained using AdamW~\cite{adamw} optimizer, with a learning rate of 1e-4 and a weight decay of 0.1. The learning rate is adjusted with a cosine scheduler. The weight decay for all one-dimensional parameters is set to 0 according to \cite{van2017l2}.
A 2-epoch linear warmup with an initial learning rate multiplier of 1e-3 is applied to stabilize training.
Drop path~\cite{larsson2016fractalnet} is applied to -L and -g variants with rates of 0.1 and 0.4, respectively.
For full fine-tuning settings, we use layer-wise $lr$ decay values~\cite{beit} of 0.7/0.8/0.9 for -B/-L/-g variants respectively.
The models are trained with automatic mixed precision enabled. The CUDA kernel optimization techniques brought by PyTorch, including torch.compile and memory-efficient attention~\cite{memory_efficient_attn}, are also used to reduce GPU memory usage and training time.

\noindent
{\bf Inference.} For all performance and efficiency evaluation experiments, automatic mixed precision and memory-efficient attention~\cite{memory_efficient_attn} are enabled.

\begin{table}[!t]
\caption{Hyper-parameters used in our models.}\label{tbl:hyperparameters}
\centering
\begin{tabular}{ l | c }
\toprule
Item & Value
\\
\midrule
\multicolumn{2}{@{}l}{\emph{overall:}} \\
\midrule
template area factor & 2 \\
search region area factor & 4 (-224) / 5 (-378) \\
horizontal flip & 0.5 \\
color jitter & 0.4 \\
batch size & 128 \\
epochs & 170 / 100 (GOT-10k) \\
optimizer & AdamW \\
lr & 1e-4 \\
weight decay & 0.1 \\
drop path & 0.0 (B) / 0.1 (L) / 0.4 (g) \\
clip max norm & 1.0 \\
warmup epochs & 2 \\
warmup lr mult & 1e-3 \\
BCE loss coef & 1.0 \\
GIoU loss coef & 1.0 \\
hann window panelty & 0.45 \\
\midrule
\multicolumn{2}{@{}l}{\emph{LoRA:}} \\
\midrule
$r$ & 64 \\
\midrule
\multicolumn{2}{@{}l}{\emph{full fine-tune:}} \\
\midrule
layer-wise lr decay  & 0.7 (B) / 0.8 (L) / 0.9 (g) \\
\bottomrule
\end{tabular}
\end{table}

\subsection{Head Network}
\noindent
{\bf Classification Branch.}
The classification branch outputs a classification response map $\mathbf{R} \in [0, 1]^{ N_S^H \times N_S^W }$, which assigns a classification score to each point on the feature map. The classification scores at foreground points, those intersecting with the target object, are set to reflect the Intersection over Union (IoU) between the bounding boxes predicted by the regression branch and the ground truth. For background points, the score is targeted to zero. This method of scoring, known as \emph{IoU-aware classification score} (IACS) as referenced in \cite{varifocal}, provides a joint assessment of object presence confidence and localization accuracy.

\noindent
{\bf Bounding Box Regression Branch.}
Concurrently, the bounding box regression branch generates a map $\mathbf{B} \in [0, 1]^{N_S^H \times N_S^W \times 4}$, predicting the bounding box coordinates relative to each point on the feature map. At each point $(x_i, y_j)$, the branch predicts offsets $(l_{i, j}, t_{i, j}, r_{i, j}, b_{i, j})$, representing the left, top, right, and bottom distances from the point to the edges of the bounding box. The predicted coordinates are calculated as $(x_i - l_{i, j}, y_j - t_{i, j}, x_i + r_{i, j}, y_j + b_{i, j})$. The point with the highest classification score is selected as the final bounding box output.

\subsection{Loss Functions}

\noindent
{\bf Classification Loss.} We adopt the binary cross-entropy (BCE) loss for classification, formulated as:
\begin{equation}
\begin{split}
\mathcal{L}_{cls} &= \frac{1}{N_{\text{pos}}} \sum_{i,j} - ( \Bar{y}_{i,j} \log(\hat{y}_{i,j}) + (1 - \Bar{y}_{i,j}) \log(1 - \hat{y}_{i,j}) ) , \\
{\textrm{where}}\ \Bar{y} &= \left\{
    \begin{array}{ll}
        \mathrm{IoU}(B_{i,j}^{\text{pred}}, B^{\text{gt}})  &  y_{i,j}=1, \\
        0 & y_{i,j}=0,
    \end{array}\right.
\end{split}
\end{equation}
where $y_{i,j}$ represents the ground-truth label of the location $(i,j)$ on the classification response map $\mathbf{R}$ (1 for foreground, 0 for background), $\hat{y}_{i,j}$ represents the \emph{IoU-aware classification score}~\cite{varifocal}, $B_{i,j}^{\text{pred}}$ is the target object bounding box predicted by the bounding box regression branch, and $B^{\text{gt}}$ is the ground-truth bounding box, $N_{\text{pos}}$ is the total number of positive samples on the response map.

\noindent
{\bf Bounding Box Regression Loss.} We adopt the GIoU~\cite{giou} loss for bounding box regression, formulated as:
\begin{equation}
\mathcal{L}_{box} = \frac{1}{N_{\text{pos}}} \sum_{i,j} \mathbbm{1}_{{y_{i,j}=1}} \text{GIoU}(B^{\text{gt}}, B_{i,j}^{\text{pred}})
\end{equation}
where $\mathbbm{1}_{{y_{i,j}=1}}$ is the indicator function, ensuring that only positive samples contribute to the loss.

The overall loss function is the sum of the classification loss and the bounding box regression loss, with equal weights (1.0) assigned to both loss terms. 

\section{Results on Additional Benchmarks}

In this section, we report the performance of the tracker on three additional benchmarks: OTB100~\cite{otb}, NFS~\cite{nfs} and UAV123~\cite{uav123}. The results are presented in Table~\ref{tbl:more_results}.

\begin{table}[t] \footnotesize
\caption{Performance of our tracker on three additional benchmarks. Bold indicates the best results.}\label{tbl:more_results}
\centering
\begin{tabular}{ c | c | c | c | c | c | c }
\toprule
&
\shortstack{LoRAT \\ -B-224 } & \shortstack{LoRAT \\ -B-378 } & \shortstack{LoRAT \\ -L-224 } & \shortstack{LoRAT \\ -L-378 } & \shortstack{LoRAT \\ -g-224 } & \shortstack{LoRAT \\ -g-378 } \\
\midrule
OTB100~\cite{otb}              &
71.4 & 71.7 & 72.3 & 72.0 & \textbf{72.6} & \textbf{72.6} \\
NFS~\cite{nfs}                 &
64.0 & 66.6 & 66.0 & 66.7 & 65.0 & \textbf{68.1} \\
UAV123~\cite{uav123}                 &
72.6 & 71.9 & 73.3 & 72.5 & \textbf{74.2} & 73.9 \\
\bottomrule
\end{tabular}
\end{table}

\section{More Ablation Studies}

\begin{table}[!t]
\caption{Ablation on the impact of ViT backbone with various pre-training methods.}\label{tbl:full_exp}
\centering
\begin{tabular}{ c | c | cc | cc | cc | c }
\toprule
\multirow{2}{*}{} &
\multirow{2}{*}{Variant} & 
\multicolumn{2}{c|}{LaSOT \cite{lasot}} &
\multicolumn{2}{c|}{LaSOT$_{\text{ext}}$ \cite{lasot_ext}} &
\multicolumn{2}{c|}{TNL2K \cite{tnl2k}} &
\multirow{2}{*}{\tabincell{c}{Speed \\ fps}}
\\
& &
{\small SUC} & {\small P} &
{\small SUC} & {\small P} &
{\small SUC} & {\small P} &
\\
\midrule
\multirow{12}{*}{\rotatebox[origin=c]{90}{LoRA}} &
\multicolumn{8}{@{}l}{\emph{MAE}~\cite{mae} \emph{pre-training:}} \\
& B-256                           & 
69.2 & 73.9 & 49.6 & 56.4 & 55.9 & 57.6 & 215 \\
& L-256                           &
72.2 & 78.8 & 50.4 & 57.5 & 59.7 & 63.3 & 125 \\
\cline{2-9}
&
\multicolumn{8}{@{}l}{\emph{CLIP}~\cite{clip} \emph{pre-training:}} \\
& B-256                           &
69.2 & 73.9 & 49.3 & 55.4 & 56.5 & 58.0 & 203 \\ 
& L-224                           &
70.3 & 75.5 & 50.3 & 57.1 & 57.2 & 59.3 & 112 \\
\cline{2-9}
&
\multicolumn{8}{@{}l}{\emph{EVA-02}~\cite{eva02} \emph{pre-training:}} \\
& B-224                           &
70.4 & 76.0 & 49.2 & 55.7 & 57.9 & 60.3 & 66 \\
& L-224                           &
72.6 & 79.2 & 52.9 & 60.7 & 60.7 & 64.3 & 44 \\
\cline{2-9}
&
\multicolumn{8}{@{}l}{\emph{DINOv2}~\cite{dinov2} \emph{pre-training:}} \\
& B-224                           &
71.7 & 77.3 & 50.3 & 57.1 & 58.8 & 61.3 & 209
\\
& L-224                           &
74.2 & 80.9 & 52.8 & 60.0 & 61.1 & 65.1 & 119 \\
\midrule
\multirow{12}{*}{\rotatebox[origin=c]{90}{\tiny Full Fine-tuning}} &
\multicolumn{8}{@{}l}{\emph{MAE}~\cite{mae} \emph{pre-training:}} \\
& B-256                           &
70.3 & 75.3 & 49.2 & 56.0 & 56.9 & 58.6 & 215 \\
& L-256                           &
72.9 & 79.3 & 50.6 & 57.6 & 60.1 & 63.8 & 125 \\
\cline{2-9}
&
\multicolumn{8}{@{}l}{\emph{CLIP}~\cite{clip} \emph{pre-training:}} \\
& B-256                           &
70.1 & 75.0 & 48.7 & 55.0 & 56.8 & 58.3 & 203 \\
& L-224                           &
71.6 & 76.9 & 51.2 & 58.1 & 57.9 & 60.0 & 112 \\
\cline{2-9}
&
\multicolumn{8}{@{}l}{\emph{EVA-02}~\cite{eva02} \emph{pre-training:}} \\
& B-224                           &
70.7 & 76.2 & 49.8 & 56.9 & 56.8 & 58.6 & 66 \\
& L-224                           &
72.7 & 79.0 & 52.2 & 59.5 & 58.8 & 61.7 & 44 \\
\cline{2-9}
&
\multicolumn{8}{@{}l}{\emph{DINOv2}~\cite{dinov2} \emph{pre-training:}} \\
& B-224                           &
70.9 & 76.2 & 50.0 & 57.1 & 58.1 & 60.1 & 209
\\
& L-224                           &
73.0 & 79.3 & 52.6 & 59.8 & 60.8 & 64.3 & 119 \\
\bottomrule
\end{tabular}
\end{table}

\subsection{Impact of Different Pre-training Methods}

We investigate the impact of pre-training techniques on the performance of our tracker, focusing on four different ViT~\cite{vit} backbones: MAE~\cite{mae}, CLIP~\cite{clip}, EVA-02~\cite{eva02} and DINOv2~\cite{dinov2}. A compact version is already provided in Table~\ref{tbl:vit_pretrain}. The resolutions of template and search region are adjusted according to the patch size of the ViT backbone. For models with patch size 14, \ie DINOv2-B and EVA-02-B, the template size is set to [112 × 112], the search region size is set to [224 × 224];  For models with patch size 16, \ie MAE-B and CLIP-B, the template and search region sizes are set to [128 × 128] and [256 × 256], respectively.

As detailed in Table~\ref{tbl:full_exp}, LoRA-based fine-tuning exhibits superior performance on larger models (the variants equipped with ViT-L size backbone). Also, LoRA-based fine-tuning works better on the models with stronger pre-training. For instance, the variants pre-trained with DINOv2 work better with LoRA-based fine-tuning across all datasets, underscoring LoRA's capability to enhance model performance by reducing catastrophic forgetting.

The self-supervised DINOv2 backbone models consistently outperform other variants under both LoRA-based and full fine-tuning configurations. The performance of the variants with EVA-02 pre-training is also competitive, demonstrating the effectiveness of its vision-language pre-training. However, the inference FPS of EVA-02 variants is significantly lower than that of other models, attributed to its several custom structures that differ from the standard ViT architecture.

\begin{table}[!t]
\caption{Performance, training time, max training GPU memory usage, inference FPS of various OSTrack variants. All OSTrack variants are implemented using official codes.}\label{tbl:full_ostrack_res}
\centering
\begin{tabular}{ c | c | cc | cc | cc | c | c | c }
\toprule
\multirow{2}{*}{Variant} & 
\multirow{2}{*}{Pre-training} & 
\multicolumn{2}{c|}{LaSOT \cite{lasot}} &
\multicolumn{2}{c|}{LaSOT$_{\text{ext}}$ \cite{lasot_ext}} &
\multicolumn{2}{c|}{TNL2K \cite{tnl2k}} &
\multirow{2}{*}{\tabincell{c}{Time \\ h}} &
\multirow{2}{*}{\tabincell{c}{Memory \\ GB}} &
\multirow{2}{*}{\tabincell{c}{Speed \\ \emph{fps}}}
\\
& &
{\small SUC} & {\small P} &
{\small SUC} & {\small P} &
{\small SUC} & {\small P} & & & 
\\
\midrule
B-256                           & MAE &
68.7 & 74.6 & 46.6 & 52.4 & 55.7 & 56.1 & 12.6 & 5.3 & 124
\\
L-256                           & MAE &
70.3 & 76.1 & 46.9 & 52.7 & 56.4 & 57.2 & 35.0 & 14.8 & 52
\\
B-224                           & DINOv2 &
66.2 & 70.7 & 45.7 & 50.8 & 53.5 & 52.2 & 11.8 & 4.8 & 122
\\
L-224                           & DINOv2 &
67.1 & 71.9 & 43.5 & 47.5 & 54.1 & 52.9 & 34.1 & 13.4 & 53
\\
\bottomrule
\end{tabular}
\end{table}

\subsection{Detailed Comparison with OSTrack}

We provide detailed performance and efficiency metrics for OSTrack with various additional backbone configurations in Table~\ref{tbl:full_ostrack_res}. Compared to our trackers, OSTrack requires significantly longer training times and higher GPU memory usage during training while achieving lower performance.

\begin{table}[!t]
\caption{Experiments on LoRA initialization methods. The base model is LoRAT-L-224.}\label{tbl:lora_init}
\centering
\begin{tabular}{ c | ccc | ccc | cc }
\toprule
 \multirow{2}{*}{Method} &
 \multicolumn{3}{c|}{LaSOT~\cite{lasot}} & \multicolumn{3}{c|}{LaSOT$_{\text{ext}}$~\cite{lasot_ext}} &
 \multicolumn{2}{c}{TNL2K~\cite{tnl2k}} \\

                                  &
{\small SUC}        & {\small P$_\mathrm{Norm}$} & {\small P}          & {\small SUC}        & {\small P$_\mathrm{Norm}$} & {\small P}          &
{\small SUC}        & {\small P}          \\
\midrule
Kaiming uniform         &
73.7           & 83.0           & 80.2           & 52.9           & 64.8           & 60.2           &
61.0           & 65.0             \\
Gaussian normal &
73.2           & 82.4           & 79.7           & 52.9           & 64.9           & 60.3           &
60.5           & 64.4             \\
Truncated normal & 
74.2           & 83.6           & 80.9           & 52.8           & 64.7           & 60.0  & 61.1 & 65.1           \\
\bottomrule
\end{tabular} 
\end{table}

\subsection{Performance under Different Initialization Methods of LoRA Weights}

Weight initialization is a crucial step for training neural networks. We compare three initialization methods: kaiming uniform (as in the official LoRA implementation), gaussian normal (as in the official paper~\cite{lora}), and truncated normal distribution (the initialization method used in BERT~\cite{bert} and ViT~\cite{vit}). The experiments are conducted on our LoRAT-L-224 variant. From the results in Table~\ref{tbl:lora_init}, we observe that the truncated normal initialization achieves the best overall performance across the three benchmarks.

\begin{table*}[!t]
\caption{Ablation comparison of LoRA \vs other PEFT methods. The base model is LoRAT-L-224 (last line). }
\label{tbl:lora_vs_other_peft}
\begin{center}
\resizebox{1.0\textwidth}{!}{
\begin{tabular}{ c | cc | cc | cc | c | c | c }
\toprule
\multirow{2}{*}{\tabincell{c}{PEFT}} &
  \multicolumn{2}{c|}{LaSOT~\cite{lasot}} &
  \multicolumn{2}{c|}{LaSOT$_{\text{ext}}$~\cite{lasot_ext}} &
  \multicolumn{2}{c|}{TNL2k~\cite{tnl2k}} &
  \multicolumn{1}{c|}{\#Params} &
  \multicolumn{1}{c|}{Trainable} &
  \multicolumn{1}{c}{Speed} \\
 & SUC  & P    & SUC  & P     & SUC  & P &
 M & \% & FPS \\
\midrule
- &
73.0 & 79.3 & 52.6 & 59.8 & 60.8 & 64.3 &
307 & 100 & 119 \\
Adapter~\cite{adapter_p} &
55.0 & 52.9 & 38.3 & 39.8 & 45.4 & 39.2 &
314 & 3.3 & 78 \\
VPT-Deep ($p=5$)~\cite{vpt} &
58.0 & 56.8    & 41.3    & 44.2      & 48.0    & 43.5 &
308 & 1.4 & 83 \\
VPT-Deep ($p=10$)~\cite{vpt} &
59.3 & 58.9 & 42.1 & 45.4 & 49.4 & 45.5 &
308 & 1.5 & 83 \\
$\mathrm{(IA)}^3$~\cite{ia3} &
59.5 & 60.0 & 43.1 & 47.5 & 49.4 & 47.0 &
308 & 1.4 & 119 \\
\midrule
$\mathrm{LoRA}_{qv}(r=4)$ &
66.2 & 70.0 & 48.5 & 54.9 & 55.7 & 57.1 &
308 & 1.5 & 119 \\
$\mathrm{LoRA}_{all}(r=16)$ &
73.0 & 79.3 & 52.9 & 60.4 & 60.4 & 64.3 &
315 & 3.6 & 119 \\
\midrule
$\mathrm{LoRA}_{qv}(r=64)$ &
71.9 & 77.7 & 52.9 & 59.2 & 59.4 & 63.0 &
314 & 3.3 & 119 \\
$\mathrm{LoRA}_{qkvo}(r=64)$ &
72.9 & 79.5 & 52.7 & 60.3 & 60.5 & 64.7 &
320 & 5.3 & 119 \\
$\mathrm{LoRA}_{mlp}(r=64)$ &
72.8 & 78.9 & 52.2 & 59.6 & 60.3 & 63.9 &
323 & 6.2 & 119 \\
$\mathrm{LoRA}_{all}(r=64)$ &
74.2 & 80.9 & 52.8 & 60.0 & 61.1 & 65.1 &
336 & 9.7 & 119
 \\
\bottomrule
 \end{tabular} }
\end{center}
\end{table*}

\subsection{Comparison with other PEFT methods}

In this section, we evaluate the effectiveness of Low-Rank Adaptation (LoRA) compared to other PEFT methods within our tracking framework. We conduct a comprehensive comparison using three representative PEFT approaches: 
Adapter~\cite{adapter}, a representative of adatper-based PEFT method~\cite{adapter, adapter_l, adapter_p}. We follow the original design as in \cite{adapter}, insert the adapter layers between the multi-head self-attention module and the subsequent residual connection; Visual Prompt Tuning (VPT)~\cite{vpt}, a representative of prompt tuning-based PEFT method for vision model; $\mathrm{(IA)}^3$~\cite{ia3}, a recent zero inference computational cost PEFT method. Furthermore, we explore various configurations of LoRA, designated as: $\mathrm{LoRA}_{qv}$: LoRA added to query and value projection matrix of ViT backbone. $\mathrm{LoRA}_{qkvo}$: LoRA added to all linear layers of multi-head self-attention modules of ViT backbone. $\mathrm{LoRA}_{mlp}$: LoRA added to all linear layers of MLP layers of ViT backbone. $\mathrm{LoRA}_{all}$: LoRA added to every linear layers of ViT backbone.

Using LoRAT-L-224 as the base model, results from Table~\ref{tbl:lora_vs_other_peft} illustrate that LoRA significantly outperforms other PEFT methods under various settings. Notably, $\mathrm{LoRA}_{qv}(r=4)$ surpasses all compared PEFT methods with only 1.5\% trainable parameters. $\mathrm{LoRA}_{all}(r=16)$ matches the performance of full fine-tuning with 3.6\% trainable parameters.
Additionally, from the bottom section of Table~\ref{tbl:lora_vs_other_peft}, we observe that applying LoRA to both MHSA and MLP obtains the best performance.

\begin{table*}[!t]\footnotesize
\caption{Ablation on attention mechanism. S-MAM: Replace the self-attention module in ViT backbone with Slimming Mixed Attention Module~\cite{mixformer_extend}. }
\label{tbl:abl_attn}
\centering
\begin{tabular}{ c | ccc | ccc | cc }
\toprule
\multirow{2}{*}{Variant} & 
  \multicolumn{3}{c|}{LaSOT~\cite{lasot}} &
  \multicolumn{3}{c|}{LaSOT$_{\text{ext}}$~\cite{lasot_ext}} &
  \multicolumn{2}{c}{TNL2k~\cite{tnl2k}} \\  
& 
{\small SUC}        & {\small P$_\mathrm{Norm}$} & {\small P}          & {\small SUC}        & {\small P$_\mathrm{Norm}$} & {\small P}          &
{\small SUC}        & {\small P}          \\ 
\midrule
\multicolumn{2}{@{}l}{\emph{LoRA:}} \\
\midrule
\multicolumn{1}{l|}{L-224} & 
74.2 & 83.6 & 80.9 & 52.8 & 64.7 & 60.0 & 61.1 & 65.1 \\
\multicolumn{1}{l|}{+ S-MAM~\cite{mixformer_extend}} &
73.5 & 82.6 & 80.1 & 52.4 & 64.1 & 59.8 & 61.0 & 64.9 \\
\midrule
\multicolumn{2}{@{}l}{\emph{Full Fine-tune:}} \\
\midrule
\multicolumn{1}{l|}{L-224} & 
73.0 & 81.8 & 79.3 & 52.6 & 64.1 & 59.8 & 60.8 & 64.3 \\
\multicolumn{1}{l|}{+ S-MAM~\cite{mixformer_extend}} &
73.0 & 82.1 & 79.6 & 52.5 & 64.0 & 59.9 & 60.0 & 63.2 \\
\bottomrule
\end{tabular}
\end{table*}

\subsection{Full self-attention \vs S-MAM}

Slimming Mixed Attention Module (S-MAM) is a variant of self-attention introduced in MixViT~\cite{mixformer_extend}, which specifically prunes the template-to-search area cross-attention. We evaluate and compare the performance of full self-attention (performing the self-attention operation on flattened template and search region tokens) and S-MAM in Table~\ref{tbl:abl_attn}.
Our findings indicate that full self-attention consistently outperforms S-MAM in both LoRA-based and full fine-tuning settings. Despite this, our tracker with S-MAM still shows improved results under LoRA-based fine-tuning compared to full fine-tuning. This demonstrates the adaptability of our model design to various one-stream tracking architectures.

\end{document}